\documentclass{article}
\usepackage[utf8]{inputenc}

\usepackage{spconf}
\ninept

\usepackage{outlines}

\usepackage{graphics}
\usepackage{epstopdf}
\usepackage{graphicx}
\usepackage{subfigure}
\usepackage{latexsym}
\usepackage{amsmath,amsfonts,amssymb,mathdots}
\usepackage{comment}
\usepackage{float}
\usepackage{mathtools}

\DeclarePairedDelimiter\floor{\lfloor}{\rfloor}

\usepackage{cite}
\usepackage{doi}

\usepackage{hyperref}
\hypersetup{
  colorlinks = true,
  allcolors = blue,
}
\usepackage[capitalise,nameinlink]{cleveref}
\usepackage{enumitem}

\usepackage{tikz}

\usepackage{xcolor}

\usepackage{graphicx}
\usepackage{adjustbox}
\usepackage{tikz}
\usetikzlibrary{arrows}
\usepackage{amsmath,amssymb,braket,tikz}
\usepackage{tikz-cd}
\usepackage{wrapfig}
 \usepackage{xcolor}
\usepackage{xcolor,colortbl}
\usepackage{lipsum}

\usepackage{booktabs}       
\usepackage{amsfonts}       
\usepackage{nicefrac}       
\usepackage{microtype}      

\usepackage{amsmath}
\usepackage{blkarray}
\usepackage{multirow}
\usepackage{tikz-cd}
\usepackage{yfonts}

\usepackage{amsthm}
\usepackage{xspace}
\usepackage{rotating}

\newtheorem{definition}{Definition}

\newtheorem{remesents the mean resultsark}[theorem]{Remark}

\definecolor{myDarkBlue}{rgb}{0.0,0.0,0.3}
\definecolor{myBlue}{rgb}{0.0,0.0,1.0}
\definecolor{myWhite}{rgb}{1.0,1.0,1.0}
\definecolor{myRed}{rgb}{1.0,0.0,0.0}
\definecolor{myDarkRed}{rgb}{0.5,0.0,0.0}
\pgfdeclareverticalshading{seismic}{60bp}{color(0bp)=(myDarkBlue); color(15bp)=(myBlue); color(30bp)=(myWhite); color(45bp)=(myRed); color(60bp)=(myDarkRed)}

\usepackage{pgfplots}
\pgfplotsset{compat=1.17}
\usepgfplotslibrary{groupplots}
\usepgfplotslibrary{fillbetween}

\pgfplotscreateplotcyclelist{mycyclelist}{%
  {red, mark=o, mark options={solid}, solid, thick},
  {blue, mark=x, mark options={solid}, dashed, thick},
  {violet, mark=*, mark options={solid}, loosely dotted, thick},
  {orange, mark=o, mark options={solid}, dashed, thick},
  {black, mark=x, mark options={solid}, solid, thick}
}
\pgfplotsset{cycle list name=mycyclelist}

\DeclareFontEncoding{LS1}{}{}
\DeclareFontSubstitution{LS1}{stix}{m}{n}
\DeclareSymbolFont{symbols2}{LS1}{stixfrak} {m} {n}
\DeclareMathSymbol{\operp}{\mathbin}{symbols2}{"A8}

\DeclareMathOperator{\drop}{DROPOUT}
\DeclareMathOperator{\simm}{sim}
\DeclareMathOperator{\meann}{mean}

\DeclareMathOperator{\prodd}{product}

\vspace{-10pt}

\title{Topo-MLP : A Simplicial Network Without Message Passing}
\name{Karthikeyan Natesan Ramamurthy$^+$, Aldo Guzm\'an-S\'aenz$^+$, Mustafa Hajij$^{*}$%
\thanks{%
MH was supported in part by the National Science Foundation (NSF, DMS-2134231).
Emails: \href{knatesa@us.ibm.com}{knatesa@us.ibm.com}, \href{aldo.guzman.saenz@ibm.com}{aldo.guzman.saenz@ibm.com}, \href{mhajij@usfca.edu}{mhajij@usfca.edu}. Authors contributed equally.%
}
}
\address{
$^+$IBM Research, USA
\\ $^*$ University of San Francisco, USA }
\date{\today}

\begin{document}

\maketitle

\begin{abstract}

Due to their ability to model meaningful higher order relations among a set of entities, higher order network models have emerged recently as a powerful alternative for graph-based network models which are only capable of modeling binary relationships. Message passing paradigm is still dominantly used to learn representations even for higher order network models. While powerful, message passing can have disadvantages during inference, particularly when the higher order connectivity information is missing or corrupted. To overcome such limitations, we propose Topo-MLP, a purely MLP-based simplicial neural network algorithm to learn the representation  of elements in a simplicial complex without explicitly relying on message passing. Our framework utilizes a novel Higher Order Neighborhood Contrastive (HONC) loss which implicitly incorporates the simplicial structure into representation learning. Our proposed model's simplicity makes it faster during inference. Moreover, we show that our model is robust when faced with missing or corrupted connectivity structure.
\end{abstract}


\section{Introduction}
\label{sec:intro}

Graphs and complex networks are prevalent and represent a cornerstone in knowledge extraction from relational data.  These characteristics enable graphs as effective tools in modeling and interfacing with a vast array of problems ranging from social networks, chemical and physical interactions, drug discovery and road networks. Although graphs can be used to model relations in data, their modeling capacity is limited to \textit{binary} relations. This results in practical challenges when modeling \textit{multiway} relationships or \textit{higher order} relationships \cite{bick2021higher}. Examples of such applications are prevalent in the real-world including higher order interactions in physical complex systems \cite{battiston2021physics} and higher order protein interactions \cite{xiao2020graph}.

Due to their prevalence in practical applications, the past few years have witnessed a surge of interest in extending deep learning protocols to higher order topological domains such as simplicial complexes \cite{roddenberry2021principled,bunch2020simplicial,schaub2020random} and hypergraphs \cite{feng2019hypergraph}. Early work on higher order topological domains focused on generalizing the graph convolutional networks to process signals on graph edges \cite{roddenberry2019hodgenet,schaub2020random}. Hypergraph convolutional operators were introduced in \cite{feng2019hypergraph}. Learning on simplicial complexes was introduced in \cite{ebli2020simplicial}, extending graph neural networks \cite{fey2019fast,morris2019weisfeiler,kipf2016semi}. Other work along these lines include \cite{roddenberry2021principled,bick2021higher,schaub2021signal,schaub2020random}.

Among the many types of higher order topological domains (e.g. hypergraphs, simplicial/cell complexes), simplicial complexes are the most well-known. This popularity stems from the their simple yet general structure \cite{hatcher2005algebraic}. Simplicial complexes form a natural generalization of many discrete domains that are important in scientific computations such as sequences, images, graphs, 3D shapes and volumetric meshes \cite{bick2021higher}. From a deep learning standpoint, higher order interactions defined on a simplicial complex represent the flow of information, and the computations that are executed between simplices. Furthermore, higher order interactions among the simplices have been empirically related to generalizability and expressiveness of neural networks -- making higher order networks defined on a simplicial complex natural objects to model these properties. Moreover, simplicial complexes also admit combinatorial structures which are specified by a collection of adjacency and/or boundary matrices. From a deep learning perspective, the combinatorial structure naturally extends to a message passing schemes deep learning paradigm \cite{hajijcell} which encompasses several state-of-the-art graph-based models \cite{fey2019fast,loukas2019graph,battaglia2018relational,morris2019weisfeiler,battaglia2016interaction,kipf2016semi} and simplicial complex-based models \cite{ebli2020simplicial,bunch2020simplicial,roddenberry2021principled}.

\begin{figure}[h]
 \centering
  {\includegraphics[scale=0.045]{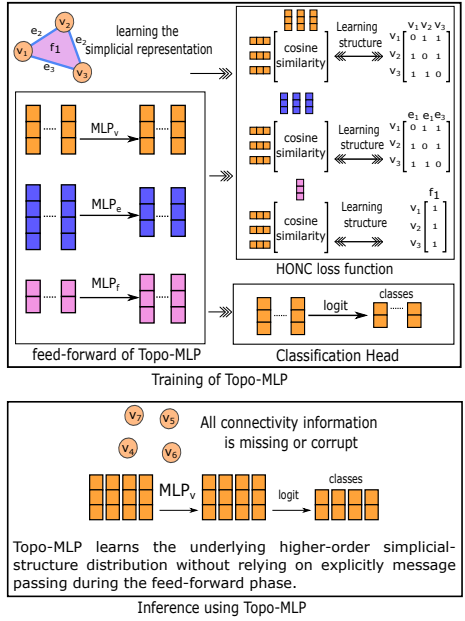}
   \caption{We learn the vertex, edge and face representations in the training phase using MLPs, while incorporating the simplicial structure using the Higher Order Neigborhood Contrastive (HONC) losses. Inference can be performed just using node features without knowledge of any connectivity information.}
 \label{main}}
\end{figure}

Our current work takes a different direction from all existing work on learning higher order structures which typically exploits the message passing paradigm to learn the simplicial structure. Specifically, we propose \textit{Topo-MLP}, a novel and pure MLP-based simplicial network that learns simplicial representation of elements in a complex without explicitly relying on message passing (see Fig. \ref{main}). Instead of imposing the structure of a simplicial complex in the model, Topo-MLP learns the representation of simplices of various dimensions by utilizing a novel Higher Order Neighborhood Contrastive (HONC) loss. This loss implicitly incorporates simplicial structure into representation learning without relying on the message passing paradigm. Because the proposed model does not rely directly on message passing, during inference Topo-MLP exhibits improved robustness to missing or corrupted underlying higher order connectivity information. Our work here extends Graph-MLP \cite{hu2021graph}, an approach for graphs which utilizes analogous notions to learn the graph structure without relying directly on graph-based message passing. 



\noindent\textbf{Contributions.}
\begin{itemize}[leftmargin=*]
\item A novel MLP-based simplicial neural network (Topo-MLP) without message passing.   
\item A novel Higher Order Neighborhood Contrastive (HONC)  loss which incorporates the simplicial complex structure into simplex representation learning.
\item Experiments that demonstrate competitive results in node classification, robustness to missing connectivity information, and faster inference times.
\end{itemize}

  \vspace{-10pt}


\section{Prior Work}

\noindent\textbf{Message Passing.}  It was shown in \cite{gilmer2017neural} that the vast majority of popular graph neural networks \cite{kipf2016semi,velickovic2017graph} can be realized as message passing models. These models are studied further in subsequent works \cite{fey2019fast,loukas2019graph,battaglia2018relational,morris2019weisfeiler,battaglia2016interaction}. On complexes, message passing was introduced in \cite{hajijcell} and studied further in \cite{schaub2021signal,schaub2020random,bick2021higher,ebli2020simplicial,hajij2022high}.

\noindent\textbf{Simplicial Networks.} Early work of learning on simplicial complexes focused on learning convolutional operators \cite{ebli2020simplicial,hajij2021simplicial,roddenberry2021principled,bunch2020simplicial}. More variations of these have appeared in \cite{keros2021dist2cycle,majhi2022dynamics,chen2021bscnets,giusti2022simplicial,hajij2022higher,yang2022simplicial}.

The fundamental difference between these and our work is that we do not depend on message passing during training/inference and we do not require connectivity information during inference (See also Section 
\ref{sec:intro}).

\section{Background}
\label{SC}
\subsection{Simplicial Complexes}

Simplicial complexes are generalization of graphs where higher order relations beyond the binary relations are allowed to occur among elements of a finite set.
\begin{definition}
An \textit{abstract simplicial complex} on a finite set $\mathcal{V}$, is a pair $(\mathcal{V},\mathcal{X})$ where $\mathcal{X}$ is subset of $\mathcal{P}( \mathcal{V} )$, the power-set of $\mathcal{V}$, such that $ x \in \mathcal{X} $ and $y \subset x $ implies $y \in \mathcal{X}$. We call $\mathcal{V}$ the set of vertices of the simplicial complex.
\end{definition}
Each simplex $s \in \mathcal{X} $ is simply an unordered subset of $\mathcal{V}$. The definition of simplicial complexes simply means that any subset of a simplex in $\mathcal{X}$ is also a simplex. The \textit{dimension} of a simplex $s$ in $\mathcal{X}$ is defined to be $|s|-1$ where $|s|$ refers to number of elements in the set $s$. A simplex $s$ of dimension $k$ will sometimes be denoted by $s^k$. The dimension of the complex $\mathcal{X}$, denoted by $\dim(\mathcal{X})$, is the maximal dimension among all its simplices. The subset of $\mathcal{X}$ of all simplices of dimension $i$ is denoted by $\mathcal{X}^i$. In particular, we refer to elements of $\mathcal{X}^0$ as the \textit{nodes} or the \textit{vertices} of $\mathcal{X}$, elements in $\mathcal{X}^1$ as the \textit{edges}, and elements of dimension 2 as \textit{faces} or \textit{triangles}. We typically denote by $v$, $e$ and $f$ to mean a vertex, an edge and a face in $\mathcal{X}$, respectively. 


To each simplex $s \in \mathcal{X}$ an \textit{orientation} is usually given, a choice of ordering on the vertices of $s$. Two orientations of $s$ are said to be equivalent if there is an even permutation that relates one to another. We denote by $\mathcal{C}_k( \mathcal{X} )$ to the $\mathbf{R}$-vector space of all linear combinations of all oriented $k$-simplices in $ \mathcal{X}$, $\mathbf{R}$ being the set of real numbers. Elements of  $\mathcal{C}_k( \mathcal{ \mathcal{X} } )$ are typically called $k$-\textit{chains}; in particular, $k$-simplices in $ \mathcal{X}$ are called \textit{primitive $k$-chains}. Typically one makes the convention that an orientation change of a primitive $k$-chain to correspond to a change in its sign; for instance, $[v_0,v_1]=- [v_1,v_0]$. Finally, if we order $ \mathcal{X}^k$ as $[s^k_0,\cdots, s^k_{|\mathcal{X}^k|-1}  ]$ then $s^k_j$ in this sequence is identified with $j^{th}$ canonical basis element in the Euclidean space $\mathbf{R}^{|\mathcal{X}^k|}$. This identification immediately provides a vector space isomophism between $\mathcal{C}_{k} ( \mathcal{X} )$ and $\mathbf{R}^{|\mathcal{X}^k| }$, which is assumed for the rest of the paper.

\noindent\textbf{Data On Simplicial Complexes.}
Feature vectors attached to simplices can be realized in terms of \textit{cochains} which are the dual elements of chains \cite{grady2010discrete,roddenberry2021principled,roddenberry2021signal}.  Specifically, we denote by $C^k( \mathcal{X}, \mathbf{R}^d )$ the linear space of functions from $X^k : \mathcal{X}^k :\to \mathbf{R}^d$. Using an ordering on $\mathcal{X}^k$, the cochain can be written as a $d-$dimensional features in $k-$simplices which we will also denote by $X^k \in \mathbf{R}^{|\mathcal{X}^k| \times d}$. Hence $0-$cochains correspond to the signals on the vertices of the graph \cite{ortega2018graph}, $1-$cochains are signals on edges and so on.


\noindent\textbf{Neighborhood and Incidence Structures of Simplicial Complexes.}
To facilitate deep learning computations on a simplicial complex, we require analogous notions to adjacency matrices typically defined on graphs and provide a notion of proximity among vertices. Unlike graphs however, simplicial complexes offer a variety of proximity notions. Each of these notions offers a tool to extend deep learning protocols to simplicial complexes. These proximity notions are summarized in so called incidence matrices as well as the higher order Laplacians which we shall define next. For more details these in the context of deep learning please see \cite{hajijcell,hajij2022higher}. 

Any finite abstract simplicial complex $X$ of dimension $n$, is completely determined by a finite sequence of matrices called the \textit{signed incidence matrices} $\{B_k \}_{k=0}^n$. Intuitively, these matrices describe the incidence between the primitive chains of $\mathcal{X}^k$ and $\mathcal{X}^{k-1}$ given a referenced orientation. See \cite{schaub2020random} for more details. In particular, the matrix $B_1$ represents the node-edge incidence matrix, whereas $B_2$ represents the edge-triangle incidence matrix. The incidence matrices can be used to define \textit{the combinatorial $k$-Hodge Laplacian} $L_k : C_{k}( \mathcal{X} ) \to C_{k}( \mathcal{X} )$ defined via $L_k = B_k^\top B_k + B_{k+1} B_{k+1}^\top$. Similarly, we can define the \textit{higher order adjacency matrix} $A_k : C_{k}( \mathcal{X} ) \to C_{k}( \mathcal{X} )$ that describes the way $k$-primitive chains are adjacent to each other. In particular $A_0$ is the regular graph adjacency matrix. Note that whereas $k$ incidence matrices provide a proximity notion among elements of $\mathcal{X}^k$ and $\mathcal{X}^{k-1} $,  higher order adjacency/Laplacian provide a proximity notion among the elements of $\mathcal{X}^k$. We will also require a proximity matrix among $\mathcal{X}^p$ and  $ \mathcal{X}^q$ . For any $p,q \in \mathbf{Z} $ with $0\leq p<q \leq \dim(\mathcal{X} ) $, the $(p,q)$-\emph{incidence matrix} $B_{p,q}$ between $ \mathcal{X}^p $ and $ \mathcal{X}^q $ of a simplicial complex is defined to be the $ |\mathcal{X}^p| \times |\mathcal{X}^q|$ matrix with $[B_{p,q}]_{ij}=1$ if $s^p_i$ is incident to $s^q_j$ and $[B_{p,q}]_{ij}=0$ otherwise\footnote{ Observe that the incidence matrix  $B_k$ is the matrix representation of \textit{the boundary map} \cite{hatcher2005algebraic} between the simplices of $X^k$ and $X^{k-1}$ given a reference orientation. On the other hand, the matrix $B_{p,q}$ describes the \textit{set-based intersection} relations between the $q$-simplices and $p$-simplices hence describing incidence similar to the hypergraph based incidence.}.

\section{Proposed Method}

Following \cite{kipf2016semi,hu2021graph}, we motivate Topo-MLP by considering a graph node classification task. We start by building a minimal convolutional simplicial network for performing node classification on a graph. We then show how to alter this network to learn the same task using our proposed approach.

\subsection{Learning on Simplicial Complexes and the Message Passing Paradigm}
\label{HOMP}


To motivate Topo-MLP, we start by building a simple convolutional simplicial network with the goal of learning a node signal while taking as input node, edge and face cochains. We use $X^{k, l} \in \mathbf{R}^{|\mathcal{X}^k|\times d_{k,l}}$ to denote the features corresponding to the $k-$cochain at layer $l$, each feature vector having dimension $d_{k,l}$. The learnable weights have the dimension $W^{k, l} \in \mathbf{R}^{ d_{k,l}\times d_{k, l+1} }$, as they take representations from layer $l$ to layer $l+1$. Considering message-passing based update of only node features ($k=0$) using node, edge, and face representations ($k=0,1,2$) from layer $l=0$ to $l=1$ \cite{hajijcell, roddenberry2019hodgenet}, we have:
\begin{equation}
\label{simple SNN}
    X^{0,1} = A_0 X^{0, 0} W^{0,0} + B_1 X^{1, 0} W^{1, 0} +
    B_{0,2} X^{2, 0} W^{2, 0}.
\end{equation} In \eqref{simple SNN} the simplicial structure is induced via $A_0$, $B_1$ and $B_{0,2}$ to update the node representations $X^{0,1}$. Such a network can be used for node-based learning tasks such as node classification. Note that we only show update for node representations as that is what is the interest of the current work. Similar approaches can be used to update representations for higher order simplices such as edges and faces.

\subsection{Topo-MLP}
Rather than injecting the simplicial structure during training as given in \eqref{simple SNN} using message passing, Topo-MLP takes a different approach to incorporate this  information during the feed-forward pass.
This is achieved by learning representations with the structure encoded with the matrices $A_0$, $B_1$ and $B_{0,2}$ using a contrastive loss. We describe our approach in three stages: (1) Feed-forward pass of Topo-MLP, where we set the architecture and the parameterization of the MLP network, (2) Higher Order Neighborhood Contrastive Loss (HONC), where  we impose the simplicial structure using a contrastive loss function, and (3) Training and Inference of Topo-MLP.

\noindent\textbf{Feed-forward Pass of Topo-MLP.}
Similar to the message passing setting in \eqref{simple SNN}, Topo-MLP takes as input three representations,  $ X^{0, 0}, X^{1, 0},$ and $X^{2, 0}$ and performs the feed-forward computations:
\begin{align} 
\label{f1}
    X^{k,1} = \drop( \sigma ( X^{k, 0} W^{k,0})), \hspace{10pt} Z^{k} =   X^{k,1} W^{k,1},  
\end{align} where $\drop$ is the standard dropout layer and $\sigma$ is the non-linear activation function. For $ 0 \leq k \leq 2 $, the embeddings $Z^{k}$  will be used for HONC loss. We also compute
\begin{equation}
\label{f3}
    Y^{0} = Z^{0} W^{0,2},
\end{equation}
and use embeddings  $Y^{0}$ with the cross-entropy loss for classification.

\noindent\textbf{Higher Order Neighborhood Contrastive (HONC) Loss.} We learn simplicial representations such that simplices neighboring each other have nearby representations in the feature space. Neighborhood information is specified via various simplicial proximity matrices. The proposed HONC loss takes the structural information encoded in the matrices $A_0$, $B_1$ and $B_{0,2}$ as well as the embeddings $Z^{k}$, $ 0 \leq k \leq 2 $, obtained during the feed-forward stage of Topo-MLP. This information is then used to construct the desired simplicial embedding without relying on explicit higher order messaging passing. 

Specifically, for each matrix $A_0$, $B_1$ and $B_{0,2}$, we construct a matrix $S_{A_0}$, $S_{B_1}$ and $S_{B_{0,2}}$ of the same size by using the embeddings $Z^{k}$, $ 0 \leq k \leq 2 $ that encode the embedding similarity as follows:
\begin{align} 
\label{sim_matrices0}
    S_{A_0} &=  \simm (Z^{0},Z^{0}),  \\  
    \label{sim_matrices1}
 S_{B_1} &=  \simm(Z^{0},Z^{1}),  \\ 
 \label{sim_matrices2}
 S_{B_{0,2}} &=  \simm (Z^{0},Z^{2}), 
\end{align} where $\simm$ is simply the cosine similarity between the rows of the two argument matrices\footnote{Note that while the embeddings $Z^{k}$, $0\leq k \leq 2$, may in general belong to feature spaces with different dimensions, this should not be an issue and a trainable matrix can be used to make the dimensions of these embeddings align. We choose to ignore this generality here and embed $Z^{k}$ in the same feature space for all $k$ for clarity of our presentation.}.

Observe that while the classical notion of dot product compares among representations of similar entities (say nodes), our notion of similarity is a generalization as we are using the notion of proximity induced by the incidence matrix to compare among representations of different entities. In other words, we think of the matrix as a proximity measure among the representations of entities that are incident. So a node $i$ is embedded in close to an edge $i$ iff they are incident. This gives a simplicial embedding where the entire simplicial structure is represented in an ambient space. The matrices given in \eqref{sim_matrices0}, \eqref{sim_matrices1}, and \eqref{sim_matrices2} along with the structure matrices $A_0$, $B_1$ and $B_{0,2}$ are then combined in the final HONC loss function. The expressions below can be used to obtain node, edge, face HONC losses respectively as:
\begin{align} 
\label{loss0}
    l_{v_j}&= -\log\left( \frac{ \sum_{i} [A_{0}]_{ij}  \exp\left([S_{A_0}]_{ij}/\mu_v\right) }{ \sum_{i} \exp\left([S_{A_0}]_{ij}/\mu_v\right) }\right)  \\  
    \label{loss1}
    l_{e_j}&= -\log\left( \frac{ \sum_{i} [B_1]_{ij}  \exp\left([S_{B_1}]_{ij}/\mu_e\right) }{ \sum_{i} \exp\left([S_{B_1}]_{ij}/\mu_e\right)}\right)  \\ 
 \label{loss2}
    l_{f_j}&= -\log\left( \frac{ \sum_{i} [B_{0,2}]_{ij}  \exp\left([S_{B_{0, 2}}]_{ij}/\mu_f\right) }{ \sum_{i} \exp\left([S_{B_{0, 2}}]_{ij}/\mu_f\right)}\right)  
\end{align} where $\mu_v, \mu_e $ and $\mu_f$ are the temperature parameters. We define the final Topo-MLP loss as:
\begin{equation*} 
\label{final}
 loss_{tot} = \beta_v \sum_{i=1}^{|\mathcal{X}^{0}|} l_{v_i} + \beta_e \sum_{j=1}^{|\mathcal{X}^{1}|} l_{e_j} + \beta_f \sum_{k=1}^{|\mathcal{X}^{2}|} l_{f_k} + loss_{CE},
\end{equation*} where the first three terms correspond to the HONC losses for vertices, edges, and faces with corresponding loss multipliers $\beta_v$, $\beta_e$ and $\beta_f$. The last term is the usual cross entropy loss used in classification.



\noindent \textbf{Training of Topo-MLP.}
Topo-MLP can be trained entirely in an end-to-end fashion. During training, the feed-forward equations given in \eqref{f1}, and \eqref{f3} do not require the simplicial information. This flexibility allows the model to be trained in batches. Specifically, to compute HONC loss for vertices we randomly sample a collection of $T_v$ vertices in $\mathcal{X}$ and take its corresponding adjacency matrix $A_0 \in \mathbf{ R }^{T_v \times T_v }  $ and node representations $X^{0, 0} \in \mathbf{R}^{T_v \times d_{0, 0}}$. Similarly, to compute HONC loss for edges, we randomly select a collection of $T_v$ nodes in $\mathcal{X}$ and $T_e$ edges in $\mathcal{X}$ and compute their corresponding boundary matrix $B_1 \in \mathbf{R}^{ T_v \times T_e  } $, the node and edge features $X^{0, 0} \in \mathbf{R}^{ T_v \times d_{0, 0} } $ and $X^{1, 0} \in \mathbf{R}^{ T_e  \times d_{1, 0} } $ respectively.  The HONC loss for faces is computed similarly\footnote{During this processing of selecting random simplices, the simplicial structure encoded in the incidence matrices is no longer maintained so when we drop a certain row, the corresponding data element is also dropped. However, we observed that this stochasticity in choosing the matrices during training improves performance experimentally.}. After training the Topo-MLP learns the simplicial embedding $Z^k$ for $0 \leq k \leq 2 $ as well as $Y^0$. Hence, Topo-MLP learns the underlying higher order simplicial structure without relying on explicitly message passing during the feed-forward phase.

\noindent \textbf{Inference using Topo-MLP.}
During inference, a message passing-based network such as the one given by \eqref{simple SNN} requires as input both the simplicial structure matrices $A_0$, $B_1$ and $B_{0,2}$ as well as the input cochains on nodes, edges, and faces. This is disadvantageous when the connectivity information of the underlying simplicial structure is missing or corrupted. On the other hand, Topo-MLP could still deliver reliable inference even when higher order connectivity information is fully missing. Note that for the node classification task the only computation that is required during inference is $  X^{0,1} = \drop( \sigma ( X^{0, 0} W^{0,0})) $  and  $Y^{0} = Z^{0} W^{0,2}$ rendering the inference computations faster in comparison to the message-passing based models.

\section{Experiments}

We apply Topo-MLP on node classification, a well-known graph data task, and demonstrate that our method applied  out-of-the-box provides competitive performance over graph-based models that are typically utilized for this task.
We demonstrate the robustness of Topo-MLP against missing
or corrupted connectivity structure. Finally, we estimate the average inference time for the proposed model and compare it against the base simplicial model given in \eqref{simple SNN}.

\subsection{Node Classification}  In our experiments, to facilitate the simplicial complex structure, we converted a graph dataset to a simplicial dataset  by converting: (1) The graph structure as well as, (2) the data supported on it to corresponding structure/data supported on a simplicial complex.

\noindent\textbf{Simplicial Data From Graphs: Structure conversion.} There are multiple ways to transform a graph to a simplicial complex. The simplicial complex that we adopt is called the \textit{clique complex} of a graph. The clique complex $\mathcal{X}_G$ of a graph $G$ is a simplicial complex obtained by considering the cliques of $G$. For our purpose we only consider the $2$-cliques complex associated with the graph $G$ \cite{bandelt2008metric}.

\noindent\textbf{Simplicial Data From Graphs: Data conversion.} To each edge $e = [v_1,v_2]$ in the $\mathcal{X}_G$ we define the data $x_e$ on it by considering an elementwise combination of the data defined on its vertices $x_{v_1}$ and $x_{v_2}$. In our experiments, the combination function $h(x_{v_1},x_{v_2})$ is chosen from among the functions $ \{ \max,\min, \meann, \prodd  \}$ based on empirical performance. The data defined on a face $f \in \mathcal{X}_G $ is defined similarly.

\noindent\textbf{Network Architecture and Training Setup.} In order to obtain a fair comparison against other reported models, we follow similar training scheme and number of parameters given in \cite{hu2021graph}. Specifically, the hidden dimension in \eqref{f1} is set to 256 for all linear layers. Dropout rate is chosen to be 0.6. We choose GELU \cite{hendrycks2016gaussian} as the activation function. For each dataset  we train for 400 epochs using Adam optimizer \cite{kingma2014adam}. Table \ref{table_results} presents the mean results of 10 independent runs with random initializations on the same dataset splits. We observe that Topo-MLP outperforms or attains similar performance against several state-of-the-art graph-based node classification models. Results for cited models are taken from the respective papers. 
  \vspace{-10pt}
\begin{table}[h!]
\footnotesize {
  \begin{center}
    \caption{Test accuracy of the Cora, Citeseer and Pubmed datasets. Best results are bolded. We report the mean of 10 independent runs with random initializations on the same dataset splits. \label{table_results} }
    \begin{tabular}{l|c|c|r} 
       & \textbf{Cora} & \textbf{Citeseer}  &\textbf{Pubmed}\\
      \hline
      DeepWalk \cite{perozzi2014deepwalk} &70.7 & 51.4 & 76.8 \\
      AdaLNet \cite{zhao2022multi} & 80.4 & 68.7 & 78.1 \\
      LNet \cite{zhao2022multi} & 79.5& 66.2 & 78.3 \\
      GCN \cite{kipf2016semi} &81.5 & 70.3 & 79.0 \\
      GAT \cite{velickovic2017graph} & \textbf{83.0} & 72.5 & 79.0 \\
      SGC \cite{wu2019simplifying} & 81.0 & 71.9 & 78.9 \\
      \hline 
      MLP & 57.8 & 54.7& 73.3 \\
      Graph-MLP \cite{hu2021graph} & 79.5 & 73.1 & 79.7 \\    
      \hline 
      Base model \eqref{simple SNN} & 79.4 & 70.3 & 79.1\\
      Topo-MLP (Ours) & 82.3\ & \textbf{73.8} & \textbf{80.9} \\  
    \end{tabular}
  \end{center}}
\end{table}

\subsection{Robustness against Corrupt or Missing Data}
\begin{figure}[!h]
  
  \centering
   {\includegraphics[scale=0.36]{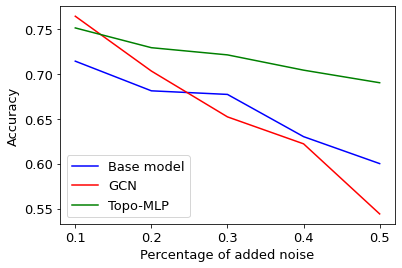}
    \caption{Adding noise to Cora dataset (training and inference). The x-axis represents the percentage noise level $\delta$ injected to the graph dataset and the y-axis represents the corresponding test accuracy. \label{fig:noise_model}}}
\end{figure}

To test the robustness of the our model against corrupt or missing data, we choose the noise model as follows. Given the graph $G=(E,V)$ and the noise ratio $\delta$, we select $ \floor{ |E| \times \delta }$ edges uniformly at random and delete them. We also add $ \floor{ |E| \times \delta } $ randomly chosen non-edges to the graph based on the original graph. We repeat this for the ratios reported in Fig. \ref{fig:noise_model} for three models with the Cora dataset, during training and inference. Observe that the GCN model is more robust to than both base model (equation \eqref{simple SNN}) and Topo-MLP when the noise ratio is relatively small. At all other noise levels, Topo-MLP dominates suggesting that the proposed method is able to learn a robust representation with better generalization. Note that increasing test noise alone for a fixed training noise level will keep the accuracy of Topo-MLP constant since it does not use the simplicial information during the test phase.

\subsection{Inference Time}
We compare the inference time of Topo-MLP against the base model (equation \eqref{simple SNN}) with a similar number of parameters. Results are reported in Table \ref{tab:running_times}. Note that inference with Topo-MLP takes $\sim 1/3$ of time taken with the base model. This is because Topo-MLP does not require the underlying simplicial structure at the inference time and only \eqref{f1} and \eqref{f3} are computed for $k=0$ which requires in our case $2$ matrix multiplies. On the other hand the base model in \eqref{simple SNN} requires $6$ matrix multiplies not counting the classification head. 

\begin{table}[h!]

\footnotesize {
  \begin{center}
    \caption{Inference time of Topo-MLP in comparison to the base model given in \eqref{simple SNN}. \label{tab:running_times}}
    \label{tab:table1}
    \begin{tabular}{l|c|c|r} 
       & \textbf{Cora} & \textbf{Citeseer}  &\textbf{Pubmed}\\
      \hline
      Base model \eqref{simple SNN} & 0.048 $\pm$ 0.0044 &  0.10 $\pm$  0.010  & 0.178 $\pm$ 0.015 \\
      Topo-MLP (Ours) & 0.0135 $\pm$ 0.003    & 0.030 $\pm$  0.005 & 0.064 $\pm$ 0.003    \\  
    \end{tabular}
  \end{center}
  }
\end{table}

\newpage
\bibliographystyle{ieeetr}
\bibliography{refs}

\end{document}